\documentclass[twocolumn]{svjour3}          
\smartqed  
\usepackage{graphicx}
\usepackage{amsmath}
\usepackage{comment}
\usepackage{booktabs}
\usepackage{color}
\usepackage{multirow}

\usepackage{hyperref}
\hypersetup{
    colorlinks=true,
    linkcolor=black,      
    urlcolor=black,
}

\begin{document}\sloppy

\title{Graph-Based Generative Representation Learning of Semantically and Behaviorally Augmented Floorplans}

\author{Vahid Azizi \and
        Muhammad Usman \and
        Honglu Zhou \and
        Petros Faloutsos \and
        Mubbasir Kapadia
}

\institute{Vahid Azizi \at
		Rutgers University, Computer Science, Piscataway, NJ, USA \\
		\email{vahid.azizi@cs.rutgers.edu}
		\and
		Muhammad Usman \at
		York University, Electrical Engineering and Computer Science, Toronto, Canada \\
		\email{usman@cse.yorku.ca}
		\and
		Honglu Zhou \at
		Rutgers University, Computer Science, Piscataway, NJ, USA \\
		\email{hz289@cs.rutgers.edu}
		\and
		Petros Faloutsos \at
		York University, Electrical Engineering and Computer Science, Toronto, Canada \\
		UHN: Toronto Rehabilitation Institute, Toronto, Canada \\
		\email{pfal@eecs.yorku.ca}
		\and
		Mubbasir Kapadia \at
		Rutgers University, Computer Science, Piscataway, NJ, USA \\
		\email{mk1353@cs.rutgers.edu}
}


\maketitle

\begin{abstract}
Floorplans are  commonly used to represent the layout of buildings. In computer aided-design (CAD)  floorplans are usually represented in the form of hierarchical graph structures.  Research works towards computational techniques that facilitate the design process, such as automated analysis and optimization, often use simple floorplan representations that ignore the semantics of the space  and do not take into account usage related analytics.

We present a floorplan embedding technique that uses an attributed graph to represent the geometric information as well as design semantics and behavioral features of the inhabitants as node and edge attributes. A Long Short-Term Memory (LSTM) Variational Autoencoder (VAE) architecture is proposed and trained to embed attributed graphs as vectors in a continuous space. A user study is conducted to evaluate the coupling of similar floorplans retrieved from the embedding space with respect to a given input (e.g., design layout). The qualitative, quantitative and user-study evaluations show that our embedding framework produces meaningful and accurate vector representations for floorplans. In addition, our proposed model is a generative model. We studied and showcased its effectiveness for generating new floorplans. We also release the dataset that we have constructed and which, for each floorplan, includes the design semantics attributes as well as simulation generated human behavioral features for further study in the community.






\end{abstract}

\section{Introduction}

Floorplan representations support a set of fundamental activities in the architectural design process, such as the ideation and development of new designs, their analysis and evaluation with respect to any selected performance criteria, and the communication among the stakeholders. While Computer-Aided Design (CAD) and Building Information Modeling (BIM) approaches support the creation of digital building models from which floorplans can be extracted, these methods do not support the systematic representation or comparison of floorplan features, which could be derived from geometric and semantic properties, as well as more advanced performance metrics, such as space utilization and occupant behaviors~\cite{merrell2010computer,feng2016crowd}.



We propose a novel technique for floorplan representation that uses a Long Short-Term Memory (LSTM) Variational Autoencoder (VAE) with attributed graphs as intermediate representations. This method considers not only the design semantics and high-level structural characteristics but also crowd behavioral attributes of potential human-building interactions. This approach represents floorplans with numerical vectors in which design semantics and human behavioral features are encoded. These vectors facilitate providing different applications related to floorplans, such as recommendation systems, real-time evaluation of designs, fast retrieval of similar floorplans and any application which needs to cluster floorplans.
The qualitative and quantitative results show the performance of our model for generating 
representative embedding vectors
such that the considered features are encoded accurately. A user-study is also conducted to validate floorplan retrievals from embedding spaces with respect to their similarity with the input floorplans. Floorplan generation is an active area of research in computer graphics.  Recently floorplan generation methods based on machine learning have been integrated in design workflows to facilitate and enhance the design process, \cite{hu2020graph2plan,chaillou2019ai+,wu2019data}. ALthought floorplan generation is not the primary goal our approach, the proposed model is a generative one and as such it can automatically generate floor plans with desired characteristics, as demonstrated by our experiments.





\subsection{Contributions}

Our contributions can be summarized as follow: (i) a workflow to represent floorplans as attributed graphs, augmented with design semantic and crowd behavioral features generated by running crowd simulations, (ii) a novel unsupervised generative deep-learning model to learn a meaningful vector representation of floorplans using  LSTM Variational Autoencoder, (iii) generation of new floorplans using the proposed model, and (iv) a user study to evaluate the qualitative performance of our approach, and (v) provision of a publicly released dataset of floorplans of indoor environments which are augmented with semantic and behavioral features. The design semantic features are extracted by our automated tool and the human behavioral features are generated by hours of running simulation.

\section{Related work}
\begin{figure*}
	\centering
	\includegraphics[width=1.0\linewidth]{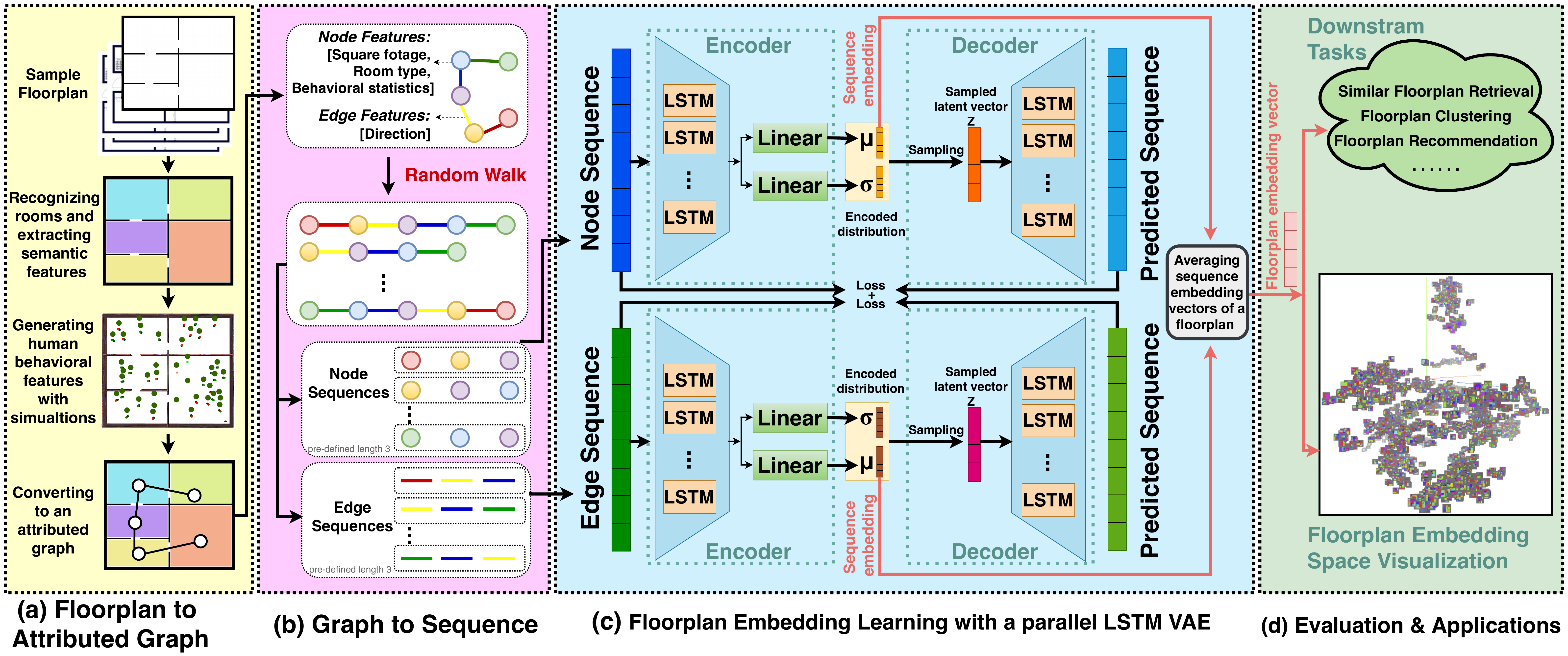} 
        \vspace{-10pt}
        \caption{\label{fig:framework} \textbf{An overview of the proposed approach.} \textbf{(a)} floorplan is firstly converted to an \textit{attributed graph} as immediate representation, with attributes residing on both nodes and edges; \textbf{(b)} \textit{random walk} is applied to the attributed graph to generate a set of node sequences and edge sequences; (\textbf{c}) floorplan embedding vector is learned with a novel parallel LSTM VAE model; (\textbf{d}) the learned floorplan embedding vector can be used for visualization and many other downstream tasks.
        }
\end{figure*}

Our work relates to research in two areas: Floorplan Representation and Floorplan Generation.

\subsection{Floorplan representation}
Floorplan representation aims for representing floorplans with numerical vectors that their structure, as well as their features, are encoded in these vectors. 
To the best of our knowledge, representing floorplans by numerical vectors are not done to date. There are some prior works for retrieving similar floorplans with representing floorplans as images or graphs. They can be mainly divided into three categories: image-based, graph-based and symbol-spotting methods.



\subsubsection{Image-based methods} 
Several approaches based on conventional image processing techniques for comparing floorplans are proposed.
In these approaches, floorplans are represented as images and Histogram of Oriented Gradients(HOG)~\cite{dalal2005histograms}, Bag of Features(BOF)~\cite{lazebnik2006beyond}, Local Binary Pattern(LBP)~\cite{ahonen2006face} and Run-Length Histogram~\cite{de2013runlength} have been utilized for extracting features from these images. Then, these extracted features are used for comparison and retrieving floorplans. By emerging Convolutional Neural Networks (CNN), In~\cite{sharma2017daniel} a deep CNN is presented for feature extraction to address the limitation of conventional image processing techniques for extracting features. These method suits object-centric floorplans datasets in which floorplans are annotated with furniture or specific visual symbols. However, these features are not semantics and do not correctly capture the high-level design structures. Moreover, human behavioral features are not considered in these methods, whereas occupants' behaviours (i.e., trajectories) are often highly correlated with environments~\cite{sohn2020laying}.


\subsubsection{Graph-based methods}
In this category, floorplans are represented with graphs and graph matching methods are utilized for measuring their similarity. Different strategies are used for representing floorplans as a graph. In \cite{sharma2016unified} rooms are nodes, and edges capture the adjacency between the rooms. In addition, nodes are augmented with furniture types annotated in floorplans. 
In~\cite{sharma2018high}, the graphs are augmented with more attributes like room area and furniture style in three different representation layers. Since in these works, floorplans are represented with graphs, all of them capture the floorplans structure, and some of them add attributes to nodes. However, mostly they do not include semantic and high-level features as well as human behavioral features. Moreover, all of these methods use graphs as a final representation and do not provide numerical vectors. These methods use graph matching methods for finding similarity, which directly applied over graphs.

\subsubsection{Symbol-Spotting methods}
Symbol spotting is a special case of Content-based Image Retrieval (CBIR) \cite{heylighen2001case,Richter2007}  which is used for document analysis. By giving a query, system retrieves zones from the documents which are likely to contain the query. Queries could be a cropped or hand-sketched image. Pattern recognition techniques are used in symbol spotting methods like moment invariants such as Zernike moments in ~\cite{lambert1995line}. Reducing search space in symbols spotting methods is proposed based on hashing of shape descriptors of graph paths (Hamiltonian paths) in ~\cite{dutta2011symbol}.
SIFT/SURF~\cite{weber2010scatch} features being efficient and scale-invariant are commonly used for spotting symbols in graphical documents. Symbol spotting methods are applied to small datasets which do not have complex images, and they are only applicable for retrieval purpose.



\subsection{Floorplan generation}
Floorplan generation aims for generating floorplan designs automatically by satisfying some constraints like room sizes and adjacency between rooms. We can divide them into two groups: Procedural\slash Optimization-based methods, and recently deep learning methods.   

\subsubsection{Procedural\slash Optimization-based methods}
In these methods, the constraints are manually defined, and optimization methods are used for constraint satisfaction to generate new floorplans. In \cite{merrell2010computer}, they used bayesian network to learn synthesizing floorplans with given high-level requirements. In \cite{rodrigues2013evolutionary,rodrigues2013evolutionary1} they proposed an enhanced Evolutionary Strategy (ES) with a Stochastic Hill Climbing (SHC) technique for floorplan generation.


\subsubsection{Deep Learning methods}
In \cite{wu2019data} a deep network was proposed for converting a given floorplan layout as input to a floorplan with predicting rooms and walls location. In \cite{chaillou2019ai+} they proposed a method comprising of three deep network models to generate floorplans. In the first step, the model generates the layout, then room locations and finally, furniture locations. In this model, users are in loop, and they can modify the input for the next steps. In \cite{hu2020graph2plan} they proposed a framework based on deep generative network. 
Users specify some properties like room count, and their model converts a layout graph, along with a building boundary, into a floorplan.  A graph-constrained generative adversarial network is proposed in \cite{nauata2020housegan}. They took an architectural constraint as a graph (i.e., the number and types) and produced a set of axis-aligned bounding boxes of rooms.


\subsection{Comparison to prior work}

Previous works do not represent floorplans with numerical vectors. They usually overlooked the design semantic and human behavioral features. In this paper, we are presenting floorplans with numerical vectors, encoded with design semantic and human behavioral features, and room directions. In addition, we also utilize generative models for addressing floorplan generation at the same time.

The proposed method is an extension of a recent approach~\cite{azizifloorplan} with the following notable extensions: 

\begin{itemize}
    \item A novel LSTM Variational Autoencoder with two branches for representing attributed graphs with features on both nodes and edges with numerical vectors. In addition, we also include rooms' directions as edge attributes to maintain layout symmetry.
    
    \item We study and showcase the generative power of our model for generating new floorplans.
    
    \item We conducted a user study to evaluate the qualitative performance of our embedding approach.
\end{itemize}

\noindent We believe these extensions significantly expand upon the previous conference paper and merit consideration for the journal. \newline








\label{sec:related-work}

\section{The proposed framework}

Fig.~\ref{fig:framework} illustrates the proposed framework, comprising of two components. The first component represents floorplans with graphs. The nodes and edges of these graphs are augmented with design semantic and human behavioral features.
The second component embeds attributed graphs in a continuous space. The details are provided in the following sections.

\subsection{Floorplan dataset}
\label{subsec:dataset}

We used the houseExpo dataset \cite{li2019houseexpo} that includes $35,357$ 2D floorplan layouts in JavaScript Object Notation format. There are 25 room types in this dataset where some of them share similar semantic labels (e.g. toilet and bathroom or terrace and balcony). We reduced the types to 10, including unknown types. This reduction is done by removing less common types based on reported statistical metrics in the dataset (e.g. freight elevator) and considering a unique type for similar propose components. The final room types are Bedroom, Bathroom, Office, Garage, Dining Room, Living Room, Kitchen, Hall, Hallway and Unknown. Unknown type is considered for room segments with a noisy label.
Additionally, we remove from the set the floorplans with inaccurate or missing labels. At the end of this process, we obtain $8,729$ floorplan layouts. This preprocessing is done to make the dataset solid for training. Corrupted data will decrease the training accuracy and, consequently, the test accuracy with undesirable outcomes of misleading the model.

\subsubsection{houseExpo++ dataset}

The original houseExpo dataset includes $35,126$ 2D floorplans. For each floorplan, the number of rooms, bounding box of the whole floorplan, a list of vertices and a dictionary of room categories, as well as their bounding boxes are provided~\cite{li2019houseexpo}. While we can use the provided bounding boxes of the rooms for segmentation, these bounding boxes are not accurate, so we use them only for labeling. We compile these floorplans (in JSON format) to images. Then, we segment the images to find the rooms, their connections if there are any, the direction of connections and their square footage. The provided bounding boxes in the original dataset are used for assigning labels to room segments by the criterion of maximum overlapping. The described processes are done with our automated tool by image processing techniques. Moreover, we convert these JSON-formated floorplans to files in a readable format with our 3D crowd simulator (SteerSuite), and by running simulations, we record the human behavior features (features are provided in Table~\ref{tab:featt}). We call this augmented dataset as \textit{houseExpo++} that is publicly available at: \url{https://github.com/VahidAz/Floorplan_dataset}. Since not all floorplans have accurate labels, we prune them and use $8,729$ floorplans for training and experiments (please refer to section~\ref{subsec:dataset} for more details).






\subsection{Floorplans to attributed graphs}
\label{sec:feature_encode}
After pruning the dataset, we represent each floorplan with a graph.
The rooms compose nodes, and the edges are their connectivity if there is an immediate door between the room pairs. For this conversion, we compiled houseExpo samples as images. Then we utilized a series of image processing techniques for room segmentation and finding their connectivity. 
Graph structure resembles the structure of floorplans like the number of rooms and their connectivity. However, considering floorplan structure is necessary but not enough. In order to have a better and more meaningful representation, we need to integrate high-level design semantic features. Moreover, humans are the inhabitant of these buildings, and their interaction with the environment provides valuable implicit information. Integration of how they interact with environments is necessary in term of safety or other type of design metrics like visibility and accessibility. Therefore, we augmented the graphs with both high-level design semantic features and human behavioral features. The nodes are augmented with both design semantic and human behavioral features and the edges only with design semantic features (Table \ref{tab:featt}).

\begin{table}[]
\setlength{\tabcolsep}{3pt}
\caption{\label{tab:featt} Features on nodes and edges.}
\begin{tabular}{|c|c|c|c|}
\hline
\multicolumn{1}{|l|}{}                                                     & \begin{tabular}[c]{@{}c@{}}Feature\\ Classes\end{tabular} & \begin{tabular}[c]{@{}c@{}}Feature\\ Types\end{tabular}                                                                                                                                                                                    & Dimension                                                                 \\ \hline
\multirow{2}{*}{\begin{tabular}[c]{@{}c@{}}Node\\ Features\end{tabular}} & \begin{tabular}[c]{@{}c@{}}Design\\ Semantic\\\hline\end{tabular} & \begin{tabular}[c]{@{}c@{}}Square Footage\\ Room types \\\hline\end{tabular}                                                                                                                                                                        & \begin{tabular}[c]{@{}c@{}}1\\ 10\\\hline\end{tabular}                            \\ 
                                                                         & Behavioral                                                & \begin{tabular}[c]{@{}c@{}}Not completed agents\\ Max evacuation time\\ Min evacuation time\\ Exit flow rate\\ Completed agents\\ Max traveled distance\\ Avg evacuation time\\ Avg traveled distance\\ Min traveled distance\end{tabular} & \begin{tabular}[c]{@{}c@{}}1\\ 1\\ 1\\ 1\\ 1\\ 1\\ 1\\ 1\\ 1\end{tabular} \\ \hline
\begin{tabular}[c]{@{}c@{}}Edge\\ Feature\end{tabular}                   & \begin{tabular}[c]{@{}c@{}}Design\\ Semantic\end{tabular} & Direction                                                                                                                                                                                                                                  & 4  \\ \hline                                                                      
\end{tabular}
\end{table}

The design semantic features include room type, square footage and the connection direction. The room types represented with a 10 dimensional one-hot vector where $roomType_i(i) = 1$ if the type is $i^{th}$ type and other entities are zero. The square footage represented with a scalar value and direction of connection with a 4 dimensional one-hot vector. We considered four main directions: North, East, South and West. Thus, $direction_i(i) = 1$ if direction belongs to $i^{th}$ direction and other entities are zero. The room types are provided as a label in dataset and both square footage and direction of connection are extracted by image processing techniques. Note that since we are not given the cardinal directions, we considered the top left corner of floorplan images as origin. Hence, the $+y$ axis points to north, and other directions are considered relatively. In addition, the direction between rooms are bidirectional and for simplicity we considered the direction from node (i.e., room) with the highest degree (room with more connections) to node with low degree. It is because usually the node (i.e., room) with the highest degree is the main room in the floorplan like the living room or hallway. The room type and square footage is specific to each room and we consider them as node features. Since the direction of the connection is a shared property between room pairs, we add it to the edge features (edge between room pairs).

\begin{figure}[hbt]
	\centering
	\includegraphics[width=0.85\linewidth]{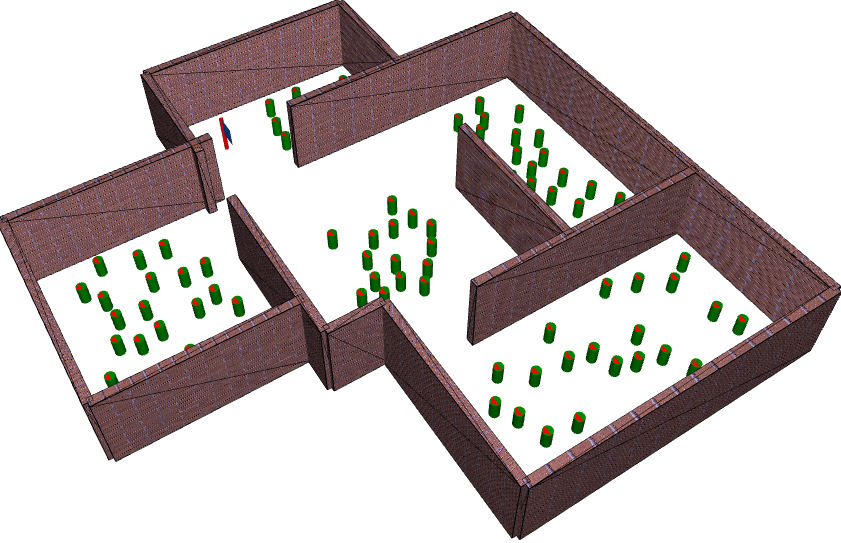} \\
    \caption{ \label{fig:sim} Crowd simulations are used to compute behavioral features for the floorplans.}
\end{figure}

The human behavioral features are generated by simulation (Fig.~\ref{fig:sim}). They include metrics regarding evacuation time, traveled distance, flow rate and the number of successful/unsuccessful agents to exit from the corresponding building (Table \ref{tab:featt}). To generate these behavioral features, we converted 2D floorplans to 3D models loadable in a crowd simulator, SteerSuite \cite{singh2009open}.  The simulator automatically populates virtual agents in each room with the target to exit the floorplan and features are calculated. All features in this class are presented with one scalar value with total dimension 9. These features are generated for each room, hence we added them to the nodes feature vector. At the end of this step, each floorplan is represented with an attributed graph $G=(V, E)$ in which $V$ denotes its vertex set (room segments) and $E \subseteq V \times V$ denotes its edges (connectivity between room pairs). Each node $v$ has a $20$ dimensional feature vector $F_{v}$ and each edge $e$ has a $4$ dimensional feature vector $F_{e}$.

\subsection{Floorplan embedding}

We convert the floorplans to attributed graphs with features on both nodes and edges (Sec.~\ref{sec:feature_encode}). These graphs represent floorplan geometry as well as their design semantics and behavioral features. They can be used directly for floorplan representation. However, graph analysis is expensive in term of computation and space cost. This challenge is addressed by proposing efficient methods for graph analysis like \cite{kumar2016g,low2012distributed,gonzalez2014graphx} but yet they are not enough efficient and these methods do not represent graphs with a compact numerical vector. Another solution for addressing the complexity of graph analysis is graph embedding. Graph embedding maps the graphs to a low dimensional space while their properties and information are maximally preserved. In this low dimensional space the graphs with similar properties are close. We have different type of graphs like heterogeneous graph, homogeneous graph, attributed graph and etc. It means the input for graph embedding methods are varying and a single method can not handle all types. Moreover, graphs can be labeled or unlabeled. Graph embedding can mainly be divided to node embedding, edge embedding, hybrid embedding and whole-graph embedding   \cite{DBLP:journals/corr/abs-1709-07604}.

In this paper, we are dealing with whole attributed graph embedding. It is because we want to represent each attributed graph (attributed floorplan) with a vector, and attributes present both on nodes and edges. These vectors encode graph (floorplan) structure as well as their design semantics and behavioral features. In addition, these graphs are unlabeled, i.e., we do not have label for each graph to perform supervised classification or regression. Moreover, the size of graph (in terms of the number of nodes) varies and is relatively small. Note that we can use other type of embedding like node embedding and then use the average (or other type of aggregation) of the node embedding vectors as the whole-graph representation. However, with this strategy the whole graph structure is not captured properly and does not lead to accurate vector representation~\cite{bai2019unsupervised,taheri2018learning}. 


There are quite a few works for the whole graph embedding.
Some whole-graph embedding methods rely on the efficient calculation of graph similarities in graph classification tasks \cite{shervashidze2011weisfeiler,niepert2016learning,mousavi2017hierarchical,bai2019unsupervised}. These methods are supervised and need a labeled dataset. In addition, they are designed for unattributed graphs. On the contrary, our graphs are attributed and unlabeled. Therefore, these types of methods are not applicable for our problem, and we need an unsupervised method. Graph2Vec~\cite{narayanan2017graph2vec} is an unsupervised method that by maximizing the likelihood of graph subtrees given graph embedding, generates vector representations. However, since this model uses subgraphs, the graph global information is not captured properly \cite{taheri2018learning}. In addition, this method is not applicable for graphs with attributes both on nodes and edges. In \cite{taheri2018learning} for capturing the whole-graph structures, they take advantage of random walk for converting graphs to a set of sequences. It is shown using random walk leads to better representation in comparison to the adjacency matrix, because the random walks capture more than the immediate neighbourhood. 
Then a LSTM autoencoder is presented to learn graph representations. However, this method suits unattributed graphs.

Sentences are presented with a sequence of words. In \cite{DBLP:journals/corr/BowmanVVDJB15,wang2019topic} LSTM Variational Autoenocder is used for text and sentence embedding and generation. The performance of converting graphs to sequences in \cite{taheri2018learning} and the methods proposed in \cite{DBLP:journals/corr/BowmanVVDJB15,wang2019topic,taheri2018learning} motivate us to convert our graphs to sequences and propose a novel LSTM Variational Autoencoder model that suits our unlabeled attributed graphs (both on nodes and edges). In particular, we convert each floorplan graph into a set of sequences (which will be described in Sec.~\ref{sec:graph2seq}), and we propose a generative model that maps our graphs (in sequences) to a d-dimensional space $\theta : G \rightarrow R^d$.
The proposed model is detailed in the next section.

\subsection{Model}
We present a novel LSTM Variational Autoencoder architecture illustrated in Fig.~\ref{fig:framework}. LSTM is a special kind of Recurrent Neural Network (RNN). It is designed for learning long-term dependencies by introducing state cell \cite{hochreiter1997long} to address the RNN problem with long term sequences \cite{bengio1994learning}. 
Autoencoders are a type of unsupervised neural network with two connected networks. The first network is an encoder that converts the inputs to latent vectors in a low dimensional space. The second network is the decoder, which reconstructs the original input vector from latent vectors \cite{kramer1991nonlinear}. However, the vanilla autoencoders map each input to a constant vector. In other words, the encoded vectors are not continuous, and interpolation is not allowed. Variational Autoencoder (VAE) is designed to address the limitation of vanilla autoencoders as a generative model. VAE is a generative model that learns the Probability Density Function (PDF) of the training data \cite{kingma2013auto}. 


As mentioned we have attributes both on nodes and edges with different dimensions. To address this difference in dimension, we consider two parallel LSTM VAE, one for node sequences and one for corresponding edge sequences. Let $S_n = {s_{n1}, s_{n2}, ..., s_{ni}}$ be a node sequence and $S_e = {s_{e1}, s_{e2}, ..., s_{ei-1}}$ be the corresponding edge sequence. We aim to learn an encoder and decoder to map between the space of these two sequences and their continuous embedding $z \in R^d$. In each branch, the encoder is defined by a variational posterior $q_\phi(z|S_n)$ and $q_\phi(z|S_e)$ and the decoder by a generative distribution
$p_\theta(S_n|z)$ and $p_\theta(S_e|z)$, where $\theta$ and $\phi$ are learned parameters. For each branch, loss function has two terms (Eq. \ref{eq:eq1}). The first term is reconstruction error that we used Mean Square Error (MSE). This term encourages the decoder to learn to reconstruct the data $\bar S_n,  \bar S_e$. The second term which is a regularizer is Kullback-Leibler divergence to penalize loss if the encoder outputs representations that are different than a standard normal distribution $N(0, 1)$ \cite{simonovsky2018graphvae}.  In training, two branch loss functions are summed for back-propagation (Eq. \ref{eq:eq1}).

\begin{multline} \label{eq:eq1}
Loss_{total} = (||S_n-\bar S_n||^2 + KL[q_\phi(z|S_n), N(0, 1)]) + \\
(||S_e-\bar S_e||^2 + KL[q_\phi(z|S_e), N(0, 1)])
\end{multline}



In both branches for the encoder we have a LSTM layer with 256 units. In the following we have two fully connected layer with dimension 16 for generating $\mu$ and $\sigma$. Then we have the sampling and finally we have the decoder with one LSTM layer with 256 units for reconstructing the input sequences. The Adam \cite{kingma2014adam} is used as optimizer. Before training, each graph is converted to a set of sequences (Sec.~\ref{sec:graph2seq}) and these sequences are used as input for training. After training, each graph is represented by averaging its sequences' embedding vectors. 



\subsection{Graphs to sequences}
\label{sec:graph2seq}
Graphs can converted to sequences by methods including but not limited to random walk or Breadth First Search (BFS). In \cite{taheri2018learning} the random walk, BFS and shortest path between all pairs of nodes are utilized. The experiments show sequences generated by random walk lead to a better vector representation. The reason is random walk captures more than immediate neighbours of nodes. Random walk is introduced in \cite{Perozzi_2014} for converting graphs to sequences. In this version, we pick a node, and then we pick one of its edges randomly to move to the next node. We repeat this procedure until we get a walk of some predefined length (length of a walk is defined by the number of nodes on the walk, and walk that is shorter than the predefined length will be padded into the predefined length). Later, two other versions are proposed.

\textit{Random Walk 1}.  In \cite{grover2016node2vec} the random walk is modified to have two parameters $Q$ and $P$. Parameter $Q$ is the probability of discovering the undiscovered parts of the graph and parameter $P$ is the probability for returning to the previous node.  We call the random walk proposed in \cite{grover2016node2vec} `Random Walk 1'.

\textit{Random Walk 2}. In \cite{taheri2018learning} the random walk is modified by adding probability $1/D(N)$, where $D(N)$ is the degree of node $N$. In this walk we start from a node and the next node will be selected by its probability $1/D(N)$. We call this random walk presented in \cite{taheri2018learning} as `Random Walk 2'.

We used both walks with different walk lengths to find the best performer walk and walk length.
Besides node sequences, the edge sequences are captured at the same time. Both of nodes and edges sequences are used for training the model.

\section{Experiment, results and evaluation}





\subsection{Training}
\label{sec:training}
As described in Sec.~\ref{sec:graph2seq}, we converted graphs to nodes and edge sequences. We utilized both mentioned walks with walk length 3, 5 and 7. For random walk 1, we set both $Q$ and $P$ to 0.5.
For each graph, we run random walk $11$ times.
Therefore, we have 11 sets of node and edge sequences for each graph. Out of 10, one of them is considered as proxy set (proxy graph). These proxy sequences are used later for evaluation.
In total we have 306440 nodes sequences and 306440 corresponding edge sequences. The sequences with length less than target length are padded with zero. We trained three models with considering different set of features.

\textit{Model 1}. In this model we only considered the design semantic features on nodes. We removed the edge branch and the model is trained only with nodes sequences. The dimension of node features is 11.

\textit{Model 2}. In this model we considered design semantic features both on nodes and edges. The model is trained with both branches. 
The features dimension on nodes is 11 and on edges is 4.

\textit{Model 3}. In this model we considered all design semantic features and human behavioral features. The model is trained with two branches. The features dimension on nodes is 20 and on edges is 4.

All three models have the same described architecture and loss function. Note that in Model 1, the edge branch is removed and we have only node branch's loss function. 
But, in other two models, the loss is summation of two branches loss. The learning rate was set empirically as 0.001. All three models are trained on a machine with 32 GB RAM, 12 * 3.50 GHz cores CPU and Quadro K620 GPU with 2 GB Memory. In average each model takes about 4 hours for training with 50 epochs.

\subsection{Quantitative evaluation}
This section presents quantitative results from the experiments.

\subsubsection{Nearest neighbours ranks}
\label{sec:nn_rank}
As mentioned, by embedding, the graphs are mapped to a continues embedding space. The graphs with similar structure and properties should be close to each other in the embedding space. The closeness of two floorplans can be measured by the euclidean distance of the two corresponding embedding vectors (smaller distance denotes higher similarity).
If this embedding space is well constructed, similar floorplans in terms of structure, semantic and behavioral properties should be close to each other.
Therefore, for each graph (called as query graph), we compute the euclidean distance from this graph to other graphs (including itself) and rank the other graphs for this graph according to the distance.
We hypothesize that each graph should find itself as the first nearest neighbour and its proxy graph (a different set of sequences for query graph) in close ranks. For this study, we use the model 3 to obtain the top 5 nearest neighbours for each floorplan in the learned embedding space. We calculate the percentage of graphs that have themselves in the first rank and the percentage of proxy graphs in the other four ranks. The table \ref{tbl:nnr} shows these percentages with different walk lengths for both random walks.

\begin{table}[]
\caption{\label{tbl:nnr} Nearest neighbour ranks (Sec.~\ref{sec:nn_rank}) with two types of random walks and walk length 3, 5 or 7. For each graph,  we  compute  the  euclidean  distance  from  this graph  to  other  graphs  (including  itself and a proxy graph which is a different set of sequences of the query graph)  and  rank  the other graphs for this graph according to the euclidean distance. Each  graph  should  find  itself  as the first nearest neighbour and its proxy graph in close ranks. We calculate the percentage of graphs that have themselves in the first rank and the percentage of proxy graphs in the other top four ranks (e.g., `[0, 75, 4, 2, 2]' in the table denotes that 75$\%$ graphs have their proxy as the top 2 nearest neighbor, 4$\%$ as  top 3, 2$\%$ as  top 4, and 2$\%$ as top 5). This analysis showcases that walk length 5 can lead to better performance, and the random walk  2 is superior. }
\begin{tabular}{|c|c|c|c|}
\hline
\multicolumn{1}{|l|}{}                                  & \begin{tabular}[c]{@{}c@{}}Walk\\ Length\end{tabular} & \begin{tabular}[c]{@{}c@{}}Rank of\\ query floorplan\\ within 5 NN\end{tabular}                               & \begin{tabular}[c]{@{}c@{}}Rank of\\ proxy graph\\ within 5 NN\end{tabular}                                  \\ \hline
\begin{tabular}[c]{@{}c@{}}Random\\ Walk 1\end{tabular} & \begin{tabular}[c]{@{}c@{}}3\\ 5\\ 7\end{tabular}     & \begin{tabular}[c]{@{}c@{}}{[}100, 0, 0, 0, 0{]}\\ {[}100, 0, 0, 0, 0{]}\\ {[}100, 0, 0, 0, 0{]}\end{tabular} & \begin{tabular}[c]{@{}c@{}}{[}0, 49, 2, 1, 1{]}\\ {[}0, 75, 4, 2, 2{]}\\ {[}0, 57, 2, 1, 1{]}\end{tabular}   \\ \hline
\begin{tabular}[c]{@{}c@{}}Random\\ Walk 2\end{tabular} & \begin{tabular}[c]{@{}c@{}}3\\ 5\\ 7\end{tabular}     & \begin{tabular}[c]{@{}c@{}}{[}100, 0, 0, 0, 0{]}\\ {[}100, 0, 0, 0, 0{]}\\ {[}100, 0, 0, 0, 0{]}\end{tabular} & \begin{tabular}[c]{@{}c@{}}{[}0 , 83, 3, 2, 1{]}\\ {[}0, 94, 2 , 1, 1{]}\\ {[}0, 89, 3, 1, 1{]}\end{tabular} \\ \hline
\end{tabular}
\end{table}



As table \ref{tbl:nnr} shows, in both random walks, walk length 5 leads to better performance. In addition, the random walk 2 is superior. By random walk 2 and walk length 5, each graph by itself is in the first rank and $94\%$ of proxy graphs in second rank. Since proxy graphs are a different set of sequences on the graphs, if the model performs properly, a good percent of the proxy graphs should be present in top ranks. Random walk 2 performs better since it captures our graphs structure better because in graphs (i.e., floorplans) we have always a main node (i.e., room) with high degree. Then moving toward this node gives the sequences that capture our graph structure better. The walk length has dependency to size of the available graphs in dataset. For us walk length 5 is the suitable length since in both random walks, the embedding performance is better in compare to walk length 3 and 5.

\subsubsection{Clustering}
As mentioned in previous section floorplans with similar properties are close in the embedding space. This similarity is in term of floorplans structure, design semantics and human behavioral features. There are many parametric method for clustering like KMeans \cite{wagstaff2001constrained} that we need to give the number of clusters as input parameter. Since we do not want to limit ourself to a specified number of clusters, we used Density Based Spatial Clustering of Applications with Noise (DBSCAN). It is a non parametric clustering method based on density. Each dense region (close packed points) represents a cluster and the points in low density regions are marked as outliers. It has two parameters, the minimum number of points in each cluster and the maximum distance between two samples in each cluster \cite{ester1996density}.

\begin{figure}[hbt]
	\centering
	\includegraphics[width=\linewidth]{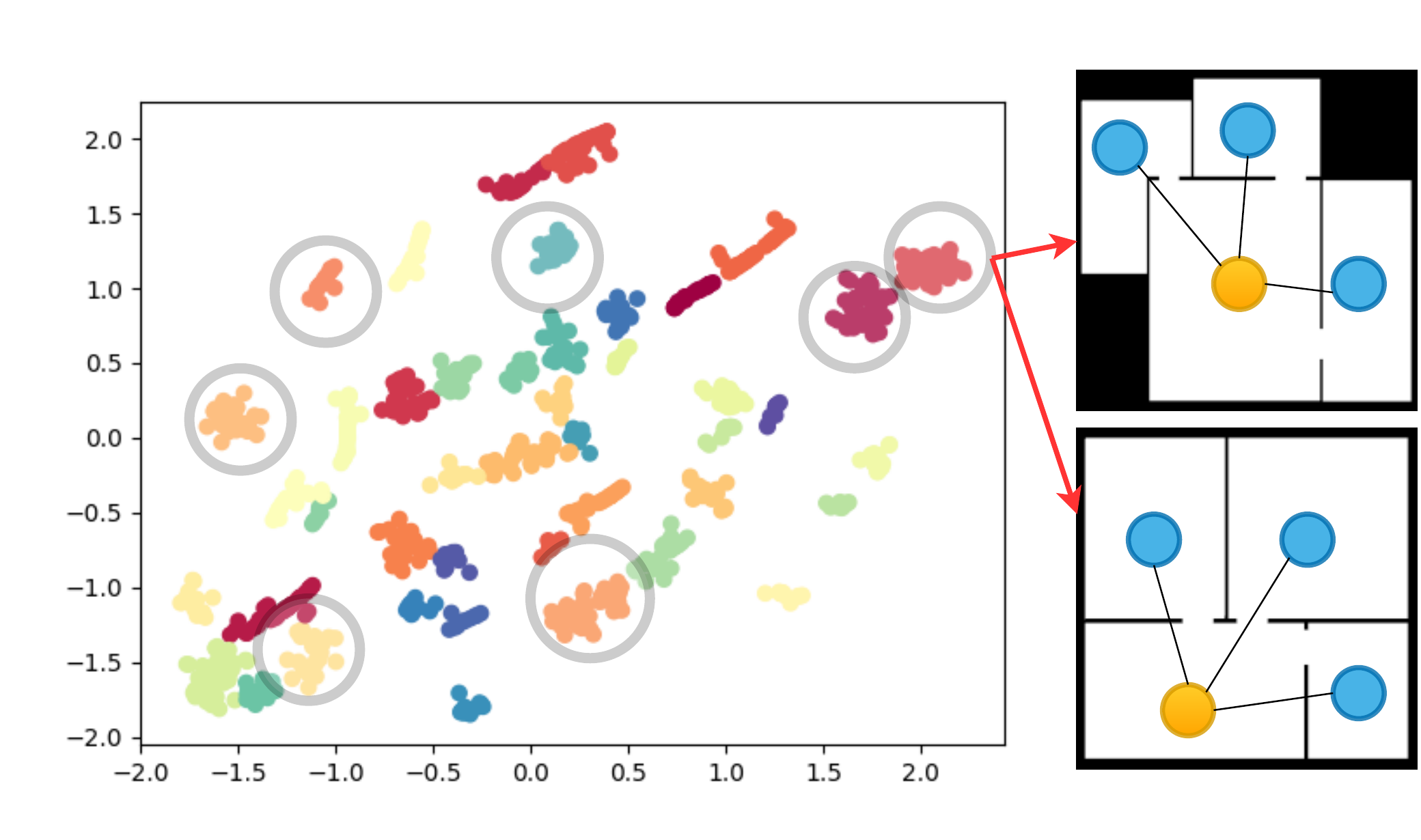} \\
    \caption{ \label{fig:clus}The clusters after running DBSCAN over 1000 random samples. 
    Two samples from one of the cluster are shown. They have the same number of nodes, same node degrees and similar room types. This shows the embedding space indeed captures the design semantics of floorplans.}
\end{figure}

We run DBSCAN over our embedding space to cluster it and then we used TSNE \cite{maaten2008visualizing} to reduce vectors dimension to two. The Fig. \ref{fig:clus} shows the resulting clusters over 1000 samples. We evaluated some clusters to see whether the samples in clusters follow the same pattern or not. We observed almost in all considered clusters the number of nodes, node degrees, node types and features are close. Except features closeness we can calculate the standard deviation for number of nodes, average of node degrees and node types for each cluster. We calculated the mentioned metrics for all clusters out of 1000 samples and their averages are provided in Table \ref{tbl:stdd}. In Fig. \ref{fig:clus}, two sample floorplans from one of the clusters depicted which shows the graph properties are encoded accurately with our model.


\begin{table}[]
\centering
\caption{Average of standard deviation for number of nodes, node degrees average and node types in clusters out of 1000 samples.} \label{tbl:stdd}
\begin{tabular}{|l|c|c|c|}
\hline
             & \textbf{\begin{tabular}[c]{@{}c@{}}Number\\ of nodes\end{tabular}} & \textbf{\begin{tabular}[c]{@{}c@{}}Average\\ of node degrees\end{tabular}} & \textbf{\begin{tabular}[c]{@{}c@{}}Node\\ types\end{tabular}} \\ \hline
\textbf{STD} & 0.16                                                               & 0.09                                                                       & 1.07                                                          \\ \hline
\end{tabular}
\end{table}

\subsection{Qualitative evaluation}
This section presents qualitative results from the experiments.

\subsubsection{Nearest neighbours (NNs) }
\label{sec:quali_nn}
\begin{figure*}
	\centering
	\includegraphics[width=\linewidth,height=\textheight]{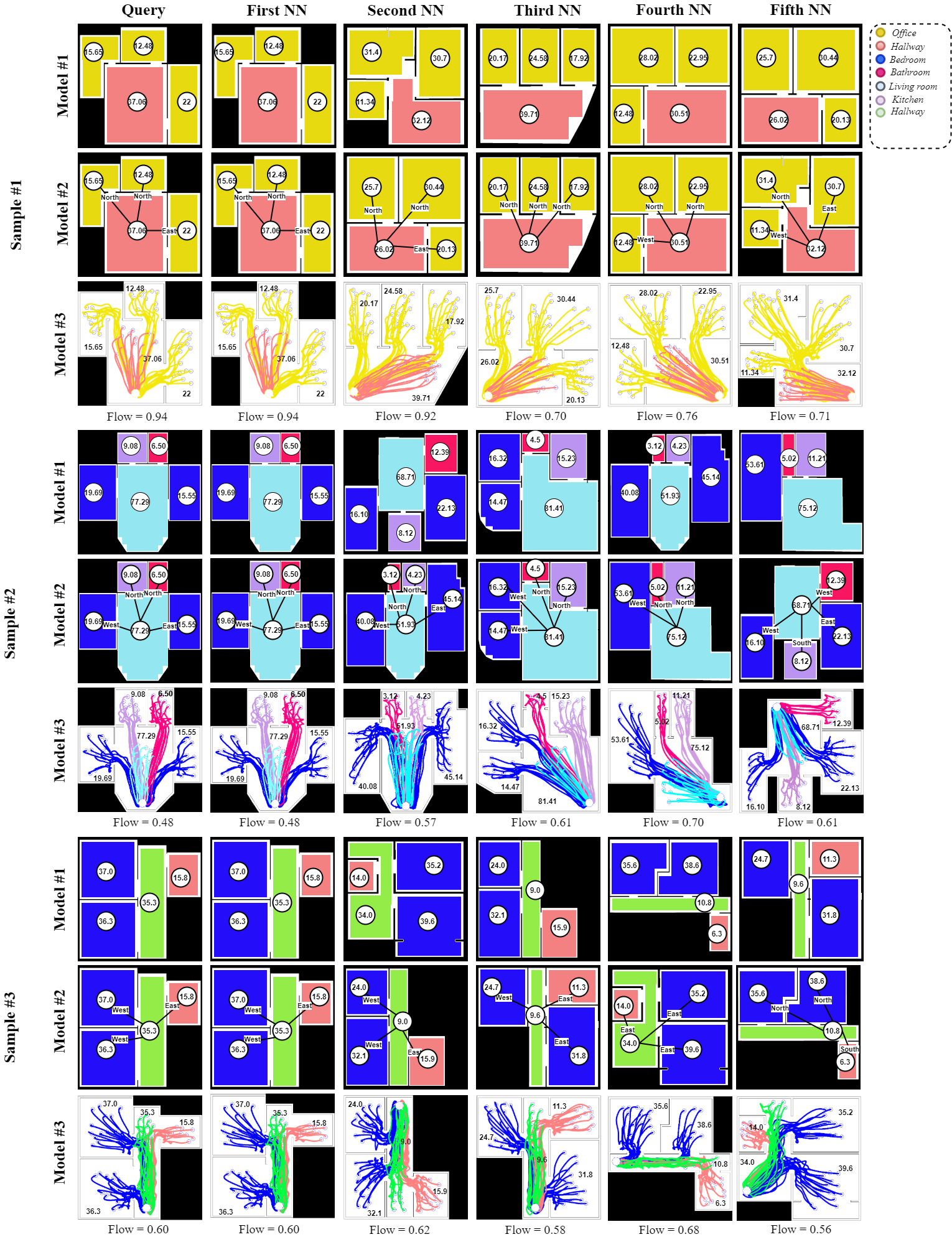} \\
	\caption{\label{fig:nnnn1}  The top 5 NNs for three floorplans found with three models. The first row shows the NNs with first model, second row shows the NNs with second model and third row shows the NNs with third model. The color of the room denotes the room type, and the numerical value inside each node denotes the square footage of the room. In the third row of each sample, the crowd flow is visualized and the numbers depicts flow rates. See Sec.~\ref{sec:quali_nn} for details.}
\label{fig:model}
\end{figure*}

As described we trained three models with different set of features. Each model makes an embedding space.  We selected three random floorplans and found their top 5 nearest neighbours in the corresponding embedding space of each model. The Fig. \ref{fig:nnnn1} shows the query floorplans and their top 5 nearest neighbours. For each sample, first row shows the NNs in the first model's embedding space, second line shows the NNs in the second model's embedding space and third row shows the NNs in the third model's embedding space. As shown in the image, with the first model, the floorplans have the same structure in term of the rooms numbers, room (node) degrees and room types. But the room arrangements are not similar. In the second model, since the edge features are added, the high rank NNs follow the same arrangement and with moving toward low rank NNs the arrangement similarity is dcreased. However, they have similar structure yet. In the third model, the human behavior features are added as well and now floorplans with similar behavioral features get close to query floorplans. The last row for each sample shows the visualization of crowd flow rate. The numbers inside floorplans in first and second row shows the square footage of each room. In third rows the numbers depict flow rates. Please note, as mentioned in section \ref{sec:feature_encode}, the north is at the top and other directions are recognized correspondingly.

\begin{figure*}
	\centering
	\includegraphics[width=0.99\linewidth]{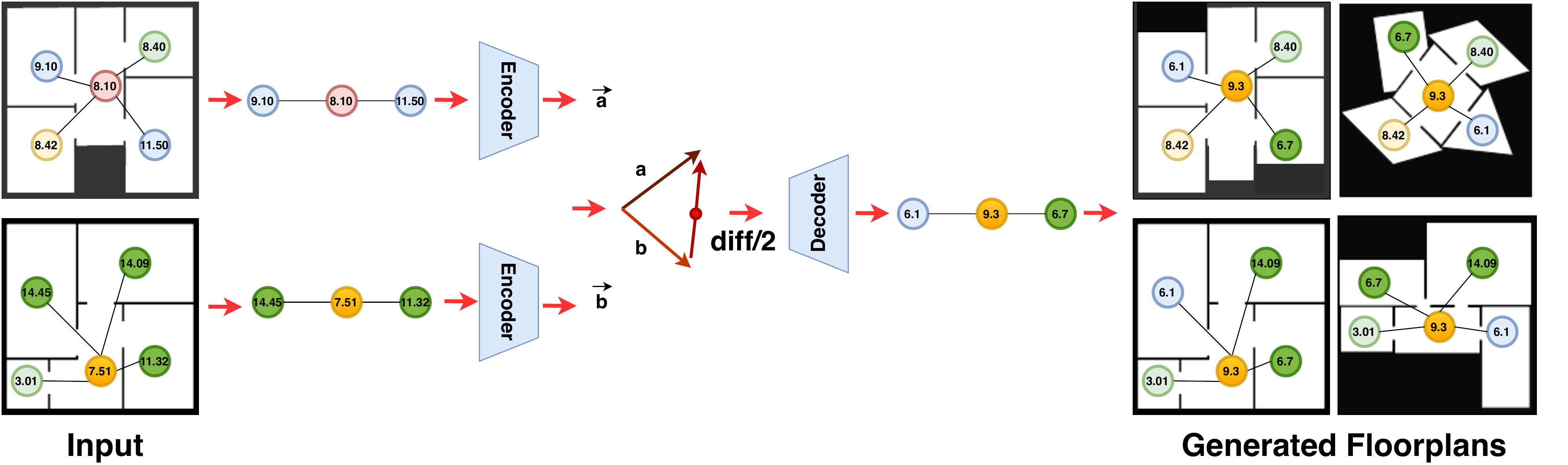} \\

\caption{\label{fig:generr} Floorplan generation with interpolating in embedding space. We select two random sequences from two random floorplans and after encoding them, we calculate the difference of their embedding vectors. Adding half of their difference to the base vector and decoding it gives us a new sequence.  These new sequence can be used for generating new floorplans.
See Sec.~\ref{sec:floorplan_generation} for details.}
\end{figure*}


\begin{figure*}[htbp]
	\centering
	\includegraphics[width=0.99\linewidth]{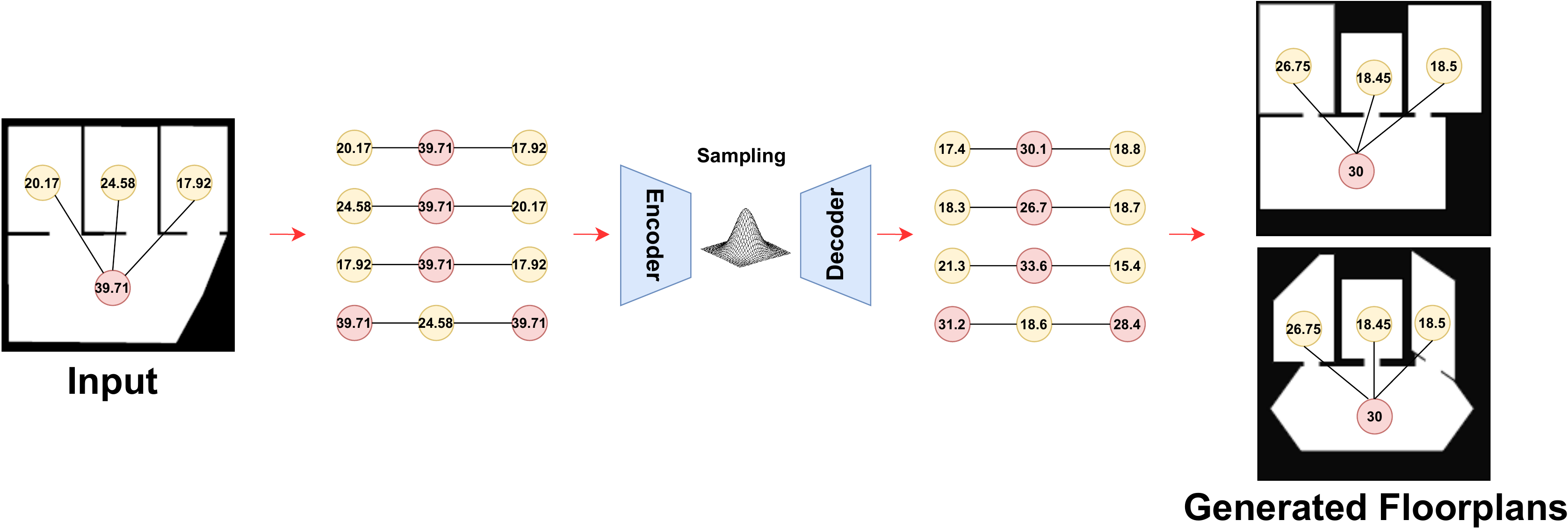} \\

\caption{\label{fig:generr_posterior} Floorplan generation with sampling from posterior distributions. To generate a new floorplan, we select a floorplan from our dataset and a set from this floorplan's 11 sets of node and edge sequences. By decoding the samples from posterior distribution of these sequences, we obtain new sequences and then map them into new floorplans. }
\end{figure*}

\section{Floorplan generation}
\label{sec:floorplan_generation}

Given that variational autoencoders are generative, we study the skill of our model for generating new floorplans. Generating new floorplans can be done with sampling from posterior distribution of sequences or with homotopies \cite{DBLP:journals/corr/BowmanVVDJB15,wang2019topic}.

\subsection{Sampling from posterior distribution}
VAE learns the data distribution instead of deterministic mapping. Therefore, we can sample from these posterior distributions for generating new data. As mentioned in Sec.~\ref{sec:training}, for each graph, we run random walk 11 times to generate 11 sets of node and edge sequences. To generate a new floorplan, we select a floorplan and a set from its 11 sets of node and edge sequences.
By decoding the samples from posterior distribution of these sequences, we get new sequences. There could be different strategies to produce a new floorplan with these newly generated sequences. We select the node with the highest degree that is repeated in all sequences as the main node. Therefore, the arrangement of other rooms can be fixed relatively. These new sequences give us information about room types and square footage. Our method does not encode the geometry shape of rooms, therefore, we can assume the room shapes are similar to the originally selected floorplan or any arbitrary shapes that satisfy the generated square footages (Fig. \ref{fig:generr_posterior}). Generating new floorplans with this approach is limited and gives us similar floorplans in terms of the number of rooms and room types, the only changes in new floorplans are the square footage and geometry shape of rooms. The square footage generated in the new sequences do not have the same value for each room, however, with considering the original sequences as references, the average of generated square footage can be used for a new floorplan.

\subsection{Homotopies}

VAE makes a continues embedding space, and it allows interpolation in this space. We used the concept of homotopy that means the set of points on the line between two embedding vectors. Instead of a set of random points, we limit our experiment to the point in the middle of the line. We can select two random sequences from two random floorplans, and after encoding them, we can calculate the difference of their embedding vectors. Adding half of their difference to the base vector and decoding it gives us a new sequence. Fig. \ref{fig:generr} shows an example.
By replacing the newly generated sequence with the old random sequence from the original floorplan,
we can generate new floorplans. It can happen between any other random sequences as well, and in this way, we can generate more derivative samples. Different strategies for interpolation and generating new floorplans could be used here. With homotopy we do not have the mentioned limitations in sampling from posterior distribution. Floorplans with varying room types can be generated as a result of interpolation, and the only limitation is the geometry shape of rooms, which is not encoded. To address this limitation, we can assume the geometry shapes are the same as reference floorplans or any arbitrary shapes that satisfy the generated square footages.

\section{User study}

\begin{table*}
	\centering
	\setlength{\tabcolsep}{8pt}
	\begin{tabular}{llll}
		\multicolumn{4}{c}{\textbf{\textit{Demographic Information}}} \\ 	\toprule
		
		\multicolumn{1}{l}{\textbf{Gender}} & \multicolumn{1}{l}{\textbf{Sex}} & \multicolumn{1}{l}{\textbf{Age}} & \multicolumn{1}{l}{\textbf{Country of Residence}} \\ \midrule
		
		{Female: 4 ($40$\%)} & {Female: 4 ($40$\%)} & {18 - 24 years old: 4 ($40$\%)} & {China: 1 ($10$\%)} \\
		
		{Male: 6 ($60$\%)} & {Male: 6 ($60$\%)} & {25 - 34 years old: 4 ($40$\%)} & {United States: 4 ($40$\%)} \\
		
		& & {35 - 44 years old: 2 ($20$\%)} & {Canada: 5 ($50$\%)} \\
		\bottomrule
		{} & {} & {} & {} \\
		
	\end{tabular}

	\setlength{\tabcolsep}{4pt}
	\centering
	\begin{tabular}{p{4.5cm}cccccp{1.5cm}}		
		\multicolumn{7}{c}{\textbf{\textit{Domain Knowledge}}} \\ 	\toprule			
		
		{} & {\textbf{Poor}} & {\textbf{Below Average}} & {\textbf{Average}} & {\textbf{Above Average}} & {\textbf{Excellent}} & \textbf{Avg. scale} \\
		\midrule 
		{Ability to interpret architectural or interior designs?} & {0 ($0$\%)} & {1 ($10$\%)} & {1 ($10$\%)} & {7 ($70$\%)} & {1 ($10$\%)} & \multicolumn{1}{c}{$3.80$} \\ 
		
		{Prior experience with architecture or interior designs?} & {2 ($20$\%)} & {0 ($0$\%)} & {1 ($10$\%)} & {6 ($60$\%)} & {1 ($10$\%)} & \multicolumn{1}{c}{$3.40$}  \\ 
		
		{Prior experience in urban planning and design?} & {3 ($30$\%)} & {0 ($0$\%)} & {2 ($20$\%)} & {5 ($50$\%)} & {0 ($0$\%)} & \multicolumn{1}{c}{$2.90$}  \\ 
		
		{Prior understanding of computational tools for architectural design-space exploration?} & {2 ($20$\%)} & {0 ($0$\%)} & {3 ($30$\%)} & {5 ($50$\%)} & {0 ($0$\%)} & \multicolumn{1}{c}{$3.10$} \\
		
		{Prior understanding of pedestrian movement flow or crowd flow?} & {0 ($0$\%)} & {2 ($20$\%)} & {4 ($40$\%)} & {3 ($30$\%)} & {1 ($10$\%)} & \multicolumn{1}{c}{$3.30$} \\				\bottomrule
	\end{tabular}

	\caption{\label{table:experts-info} Demographic information and domain knowledge ratings of expert participants (self-reported). }
\end{table*}

In this section, we present a user study to evaluate the quality and efficiency of our models of graph embeddings. Three different embedding models are tested: (1) trained with design semantic features alone, (2) design semantic and edge features, and (3) design semantic, edge and behavioral features. Given a floorplan (input), we query five similar floorplans (nearest neighbours) from each embedding model. Our hypothesis is twofold: (a) the user perceived sequence of floorplans as top-five nearest neighbours matches with the sequence captured by our model as nearest neighbours, and (b) users perform better in their perceived sequence of top-five nearest neighbours for models (2) and (3) than model (1) which is only trained with design semantic features.

\subsection{Apparatus} Floorplans are presented as 2D blueprints (e.g. a top-down skeletal view of an environment layout). The users (e.g. study participants) viewed these blueprints as high-resolution images on their own computer screens via an online survey. For model (1), each room in a floorplan is annotated with room dimension (e.g., square footage area) and color-coded with respect to its room type. The annotation for model (2) is similar to model (1) with an addition of edges between rooms and their respective directions (e.g., North, East, West, South). For model (3), we showed color-coded trajectories of virtual occupants from the rooms they spawned-in to the exit, along with square footage area of each. 

\subsection{Participants} Ten ($10$) domain experts from the architecture community ($4$ female and $6$ male) voluntarily participated in the user study. Table~\ref{table:experts-info} shows the demographic information and domain knowledge of the experts. On average, all the participants had above-average experience and expertise in interpreting architecture designs and were Knowledgeable of computational tools for design space exploration (self-reported).

\subsection{Procedure and Task} The user study is conducted as an online survey and delivered in four parts. In part (a), users are asked to provide their demographic information and report the domain knowledge and expertise in architecture and urban design. In part (b), users are presented with five different input floorplans. For each input floorplan, a sequence of 5 nearest neighbours are presented in a randomized order, which are retrieved using model (3), and presented to the users ``without'' any visual annotations. Users are asked to interactively reorder the given sequence of floorplans (e.g., via drag and drop), based on their perceived "similarity" of these floorplans with respect to input floorplan. The ordering sequence is arranged such that, more a floorplan is towards left in the order, the nearest it gets to the input floorplan. In parts (c), (d) and (e), the nearest neighbours are retrieved using models (1), (2) and (3) respectively, for the same five input floorplans which are used in part (b). In parts (c), (d), and (e), the floorplans are presented to the users ``with'' visual annotations for their respective features. We estimated that the user study will take up to $15$ minutes at maximum to complete. 

\subsection{Independent and Dependent Variables} Input floorplans and the retrieved nearest neighbours from the models are the primary independent variables. The rearranged sequences of floorplans by the users are the only dependent variables. 

\subsection{Results} 
\begin{figure*}[t]
	\centering
	
	\begin{tabular}{cc}
	
	    {{Model (0): Design Semantic + Edge + Behaviour}} & {{Model (1): Design Semantic}} \\
	    (\textit{Without Annotation}) & (\textit{With Annotation}) \\
	    
	    \includegraphics[width=0.45\linewidth]{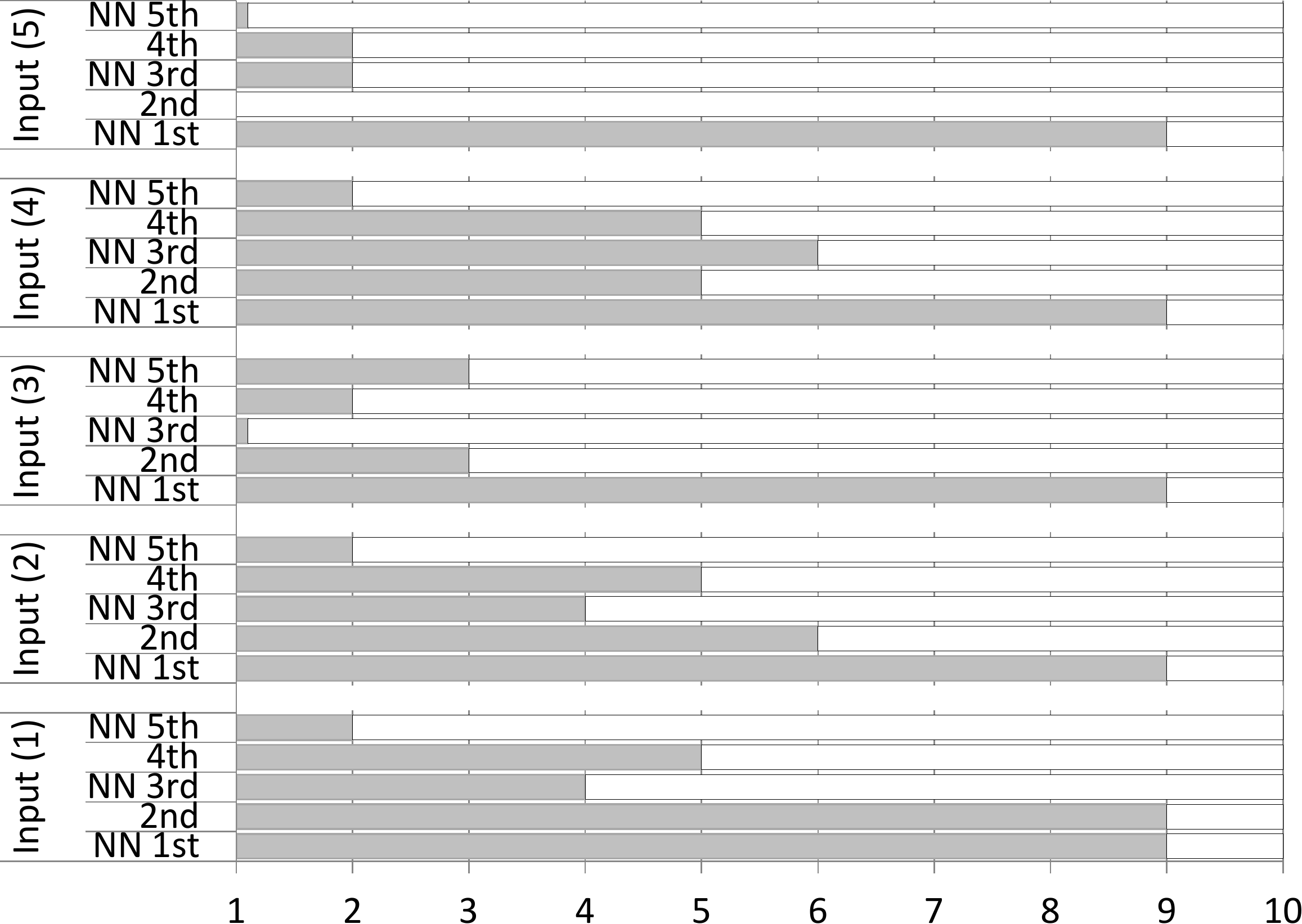} & \includegraphics[width=0.45\linewidth]{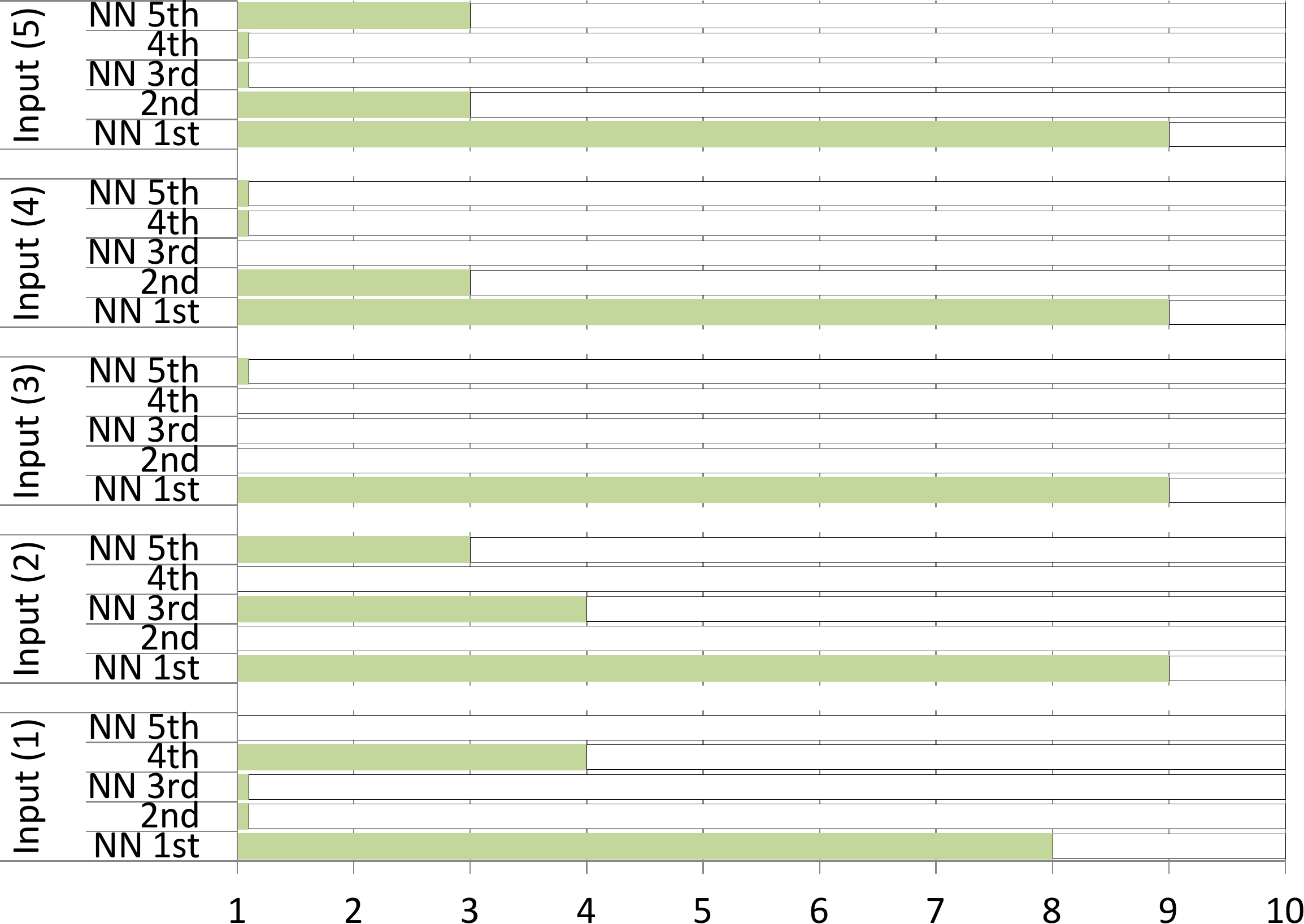} \\
	    
	    {} & {} \\
	    
	    {{Model (2): Design Semantic + Edge}} & {{Model (3): Design Semantic + Edge + Behaviour}} \\
	    (\textit{With Annotation}) & (\textit{With Annotation}) \\
	    
	    \includegraphics[width=0.45\linewidth]{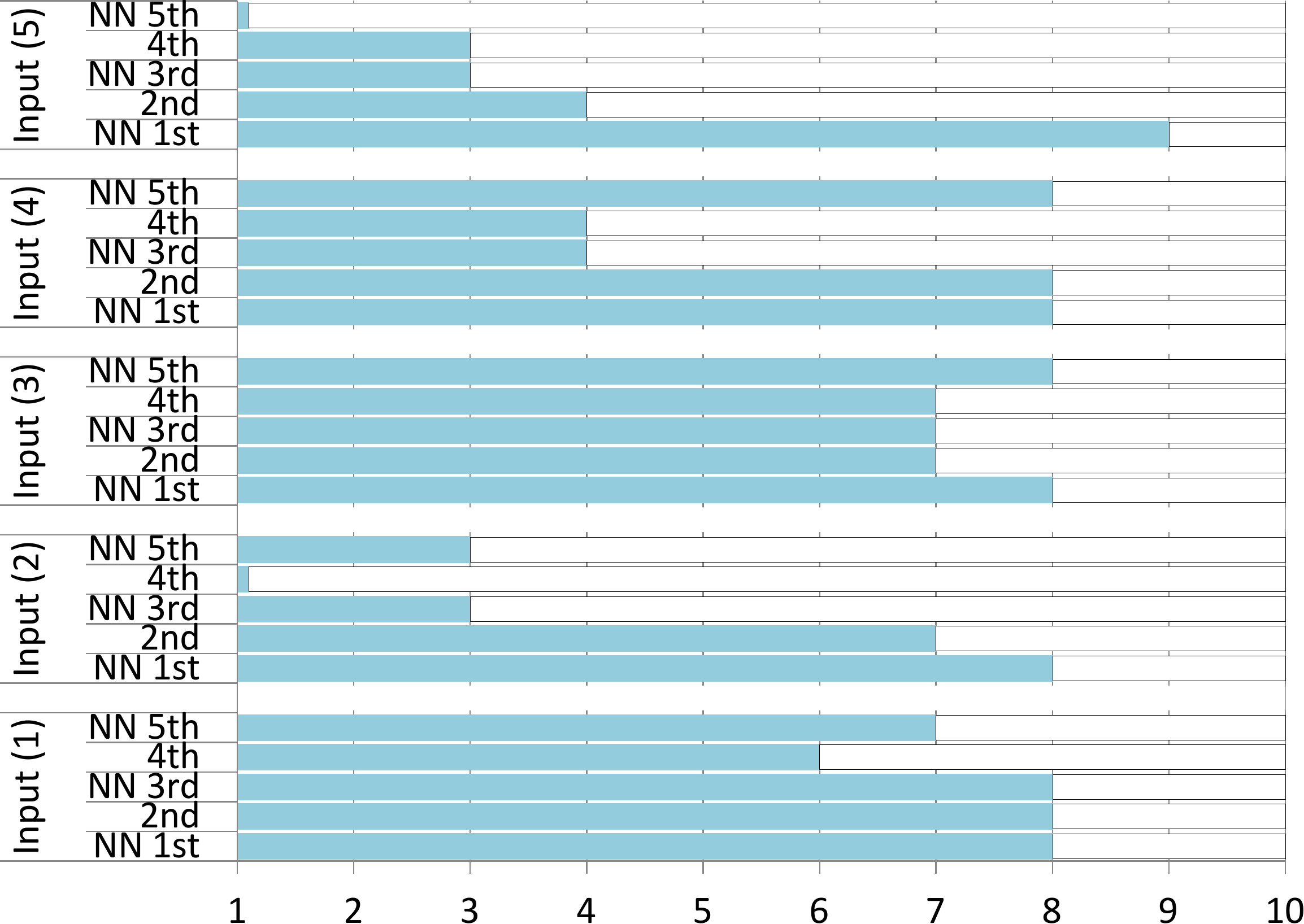} & \includegraphics[width=0.45\linewidth]{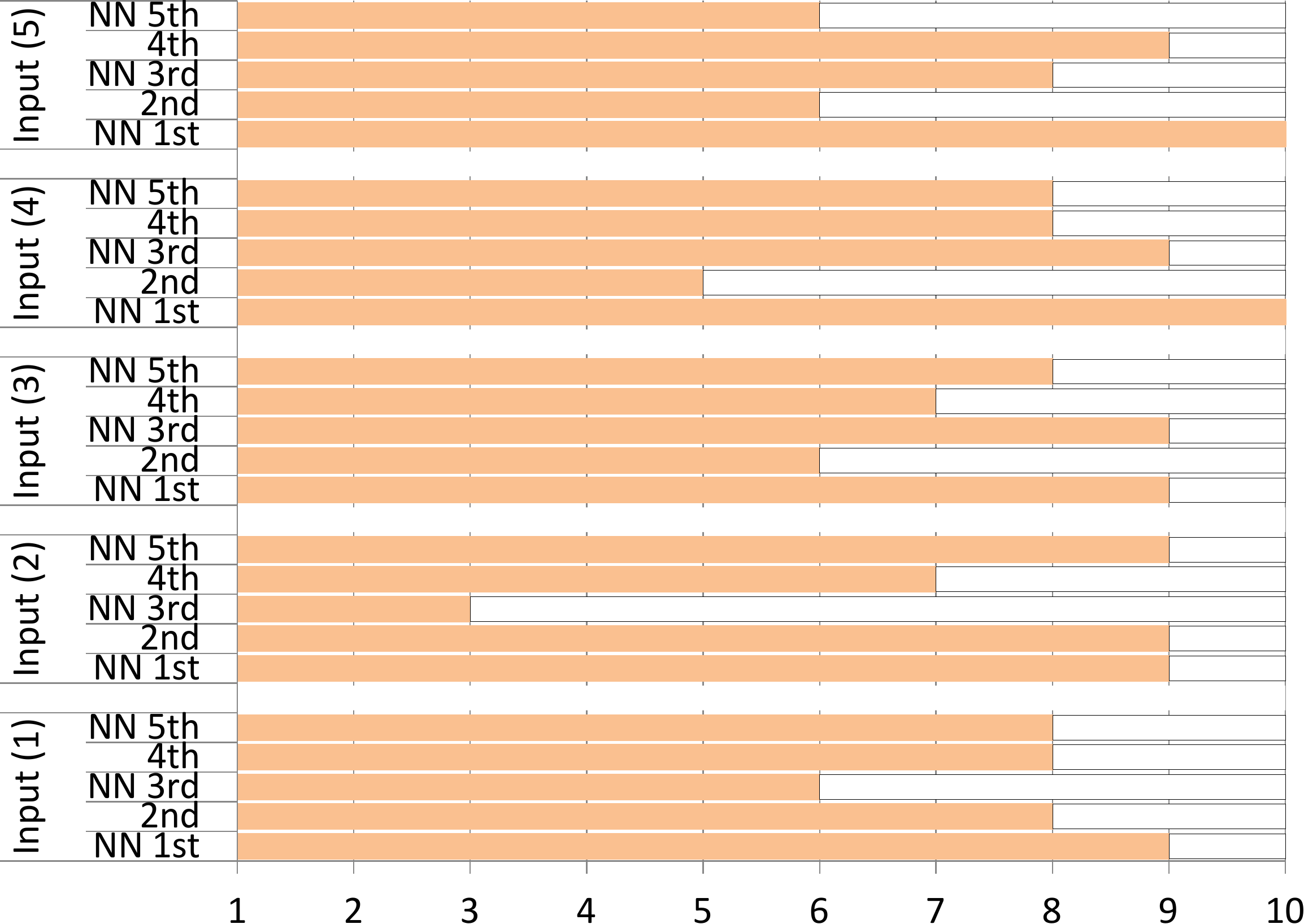} \\
	    
	    {\hspace{1.3cm} Participants} & {\hspace{1.3cm} Participants} \\
	    
    \end{tabular}
	\caption{\label{figure:user-sequences} The accuracy of user-ordered sequences of the nearest neighbours. Colored bars for each neighbour of an input floorplan represent the number of users who correctly perceived the order of the neighbour in the given sequence. Gray bars are for the nearest neighbours which are queried using model (3) but were presented to users without any annotations, Green bars are for nearest neighbours which are queried using model (1) and were presented with annotations, blue bars using model (2) --  with annotations, and orange bars using model (3) --  with annotations. }
\end{figure*}

Figure~\ref{figure:user-sequences} shows the user-ordered sequences of the nearest neighbours for the three models from user study.  The colored bars for each neighbour of an input floorplan represent the number of users who correctly perceived the order of the neighbour in the given sequence. Overall, about $28.68$\% of the neighbours are accurately ordered in their sequences based on their perceived similarity with respect to input flooplans for model (1), $59.28$\% for model (2) and about $77.6$\% for model (3), collectively by all the users. These results highlight that users least performed when they had to perceive the similarity between floorplans by considering the design semantic features alone, whereas they performed comparatively better when presented with the neighbours annotated with edge and/or behavioural attributes. The users performed the best for the model (3) when presented with the floorplans visually annotated with design semantics (e.g., room types), edge (e.g., movement direction of the agents), and behavioural (e.g., movement flow of the agents) features. The findings from the user study suggest that both of our hypothesis stand valid.

We also wanted to analyze the users' performance in perceiving the ordering sequence of the neighbours when floorplans are not visually annotated with their respective features. To test this, we used the input floorplans and their neighbours from model (3) and presented them as model (0) in the user study. These floorplans were presented to the users without any visual annotations. This was so we could analyze how important is the visual annotation of the features, and its significance to assist users in perceiving the neighbours in their correct order. Interestingly, about $ 45.68$\% of the neighbours were accurately ordered in their sequences based on their perceived similarity with respect to input flooplans for model (0). This result revealed that the annotations for design semantic features, alone, are not a good representative to convey the spatial feature information of the floorplans. As well, that the users better perceive the floorplans retrieved from the embedding space that is trained not only with the design semantic feature alone but also with the additional edge or/and dynamic behavioural features.


\section{Conclusion} 

This paper aims to represent floorplans with numerical vectors such that design semantic and human behavioral features are encoded.  Specifically,  the framework consists of two components. In the first component, an automated tool is designed for converting floorplan images to attributed graphs. The attributes are design semantic and human behavioral features generated by simulation. In the second component, we proposed a novel LSTM Variational Autoencoder for both embedding and generating floorplans. The qualitative, quantitative and expert evaluation shows our embedding framework produces meaningful and accurate vector representations for floorplans, and its abilities for generating new floorplans are showcased. In addition, we make our dataset public to facilitate the research in this domain. This dataset includes both the extracted design semantics features and simulation generated human behavioral features.

This contribution holds promise to pave the way for novel developments in automated floorplan clustering, exploration, comparison and generation. By encoding latent features in the floorplan embedding, designers can store multi-dimensional information of a building design 
to quickly identify floorplan alterations that share similar or different features. While in this work, we encode features derived from dynamic crowd simulations of building occupancy, the proposed approach can virtually scale to encode any kind of static or dynamic performance metric.

\section*{Declarations}

\subsection{Funding}
Not applicable
\subsection{Conflicts of interest} 
The authors declare that they have no conflict of interest.

\subsection{Availability of data and material (data transparency)}
The dataset is public and the reference to data is available inside paper.

\subsection{Code availability (software application or custom code)}
Not applicable

\subsection{Authors' contributions}
Not applicable

\bibliographystyle{spmpsci}      
\bibliography{main}



\end{document}